# Institutionally Distributed Deep Learning Networks

Ken Chang[1†], Niranjan Balachandar[2†], Carson K Lam[2], Darvin Yi[2], James M Brown[1], Andrew Beers[1], Bruce R Rosen[1], Daniel L Rubin[2†*], Jayashree Kalpathy-Cramer[1,3†*]

[1]Athinoula A. Martinos Center for Biomedical Imaging, Department of Radiology, Massachusetts General Hospital, Boston, MA, USA
[2]Department of Radiology and Biomedical Data Science, Stanford University, Palo Alto, CA, USA
[3]MGH and BWH Center for Clinical Data Science, Massachusetts General Hospital, Boston, MA, USA

[†]These authors contributed equally
[*]Co-Corresponding Author: Daniel Rubin, Department of Biomedical Data Science and Radiology, Stanford University, 1201 Welch Road, Stanford, CA 94305. Phone: 650-497-8716; Fax: 650-723-5795; E-mail: dlrubin@stanford.edu
Co-Corresponding Author: Jayashree Kalpathy-Cramer, Athinoula A. Martinos Center for Biomedical Imaging, 149 13th Street, Charlestown, MA 02129. Phone: 617-724-4657; Fax: 617-726-7422; E-mail: kalpathy@nmr.mgh.harvard.edu

**Abstract**

Deep learning has become a promising approach for automated medical diagnoses. When medical data samples are limited, collaboration among multiple institutions is necessary to achieve high algorithm performance. However, sharing patient data often has limitations due to technical, legal, or ethical concerns. In such cases, sharing a deep learning model is a more attractive alternative. The best method of performing such a task is unclear, however. In this study, we simulate the dissemination of learning deep learning network models across four institutions using various heuristics and compare the results with a deep learning model trained on centrally hosted patient data. The heuristics investigated include ensembling single institution models, single weight transfer, and cyclical weight transfer. We evaluated these approaches for image classification in three independent image collections (retinal fundus photos, mammography, and ImageNet). We find that cyclical weight transfer resulted in a performance (testing accuracy = 77.3%) that was closest to that of centrally hosted patient data (testing accuracy = 78.7%). We also found that there is an improvement in the performance of cyclical weight transfer heuristic with high frequency of weight transfer.

**Introduction**

With the advent of powerful graphics processing units, deep learning has brought about major breakthroughs in tasks such as image classification, speech recognition, and natural language processing.[1–3] Due to the proficiency of neural networks at pattern recognition tasks, deep learning has created practical solutions to the challenging problem of automated medical diagnoses. Recent studies have shown the potential of deep learning in diagnosing diabetic retinopathy, classifying dermatological lesions, and assessing medical records.[4–6]

Deep learning models take raw data as input and apply many layers of transformations to calculate an output signal. The high dimensionality of these transformations allows these algorithms to learn complex patterns with a high level of abstraction.[7]

A requirement for the application of deep learning within the medical domain is a large quantity of training data, especially when the difference between imaging phenotypes is subtle or if there is large heterogeneity within the population. However, patient sample sizes are often small, especially for rarer diseases.[8] Small sample sizes combined with differences in data acquisition between different institutions may result in a neural network model with low generalizability.

A possible solution to the foregoing challenges is to perform a multicenter study, which can significantly increase the sample size as well as sample diversity. Ideally, patient data is shared to a central location where the algorithm can then be trained on all the patient data. However, there are challenges to this approach. First, if the patient data takes up a large amount of storage space (such as very high-resolution images), it may be cumbersome to share these data. Second, there are often legal or ethical barriers to sharing patient data, making dispersal of some or all of the data not possible.[8] Third, patient data is valuable, so institutions might simply prefer not to share data.

In such cases, instead of sharing patient data directly, sharing the trained deep learning model may be a more appealing alternative. The model itself has much lower storage requirements than the patient data and does not contain any individually-identifiable patient information. Thus, distribution of deep learning networks across institutions can overcome the weaknesses of distributing the patient data. However, the optimal method of performing such a task has not yet, to our knowledge, been studied.

In this study, we simulate the dissemination of deep learning networks across four institutions using various heuristics and compare the results with a deep learning model trained on centrally hosted patient data. The performance of the various models is assessed via independent validation and testing cohorts. We demonstrate these simulations on 3 datasets: Kaggle Diabetic Retinopathy, Digital Database for Screening Mammography, and ImageNet.

**Methods**

*Image Preprocessing in Initial Image Collection*
We obtained 35,126 color digital retinal fundus (interior surface of the eye) images from the Kaggle Diabetic Retinopathy competition.[9] Each image was rated for disease severity by a licensed clinician on a scale of 0-4 (absent, mild, moderate, severe, and proliferative retinopathy, respectively). The images came from 17,563 patients of multiple primary care sites throughout California and elsewhere. The acquisition conditions were varied, with a range of camera models, levels of focus, and exposures. In addition, the resolutions ranged from 433x289 pixels to 5184x3456 pixels.[10] The images were pre-processed via the method detailed in the competition report by the winner, Ben Graham.[11] To summarize his method, the OpenCV python package was used to rescale images to a radius of 300, followed by local color averaging and image clipping. The images were then resized to 256x256 to reduce the memory requirements for training the neural network, while still retaining the salient features required for diagnosis. To simplify training of the network, the labels were binarized to Healthy (scale 0) and Diseased (scale 2, 3, or 4). Furthermore, mild diabetic retinopathy images (scale 1, n = 2443 images), which represent a middle ground between Healthy and Diseased, were not used for our experiments. It is also known that there is a correlation between the disease status of the left eye and the status of the right eye. To remove this as a confounding factor in our study, only images from left eye were utilized.

*Division of "Institutions", Validation, and Testing Cohorts*
The dataset was randomly sampled, with equal class distributions, into 4 "institutions", each institution having n = 1500 patients. In addition, the dataset was sampled to create a single validation cohort (n = 3000 patients) and a single testing (n = 3000 patients) cohort, again with equal class probabilities (Fig. 1B). Sampling was without replacement such that there are no overlapping patients in any of the cohorts. The image intensity was normalized within each channel across all patients within each cohort. Because model performance plateaus as the number of training patient samples increases, the number of patients per institution was limited to 1500 to prevent saturation of learning for models trained in single institutions.

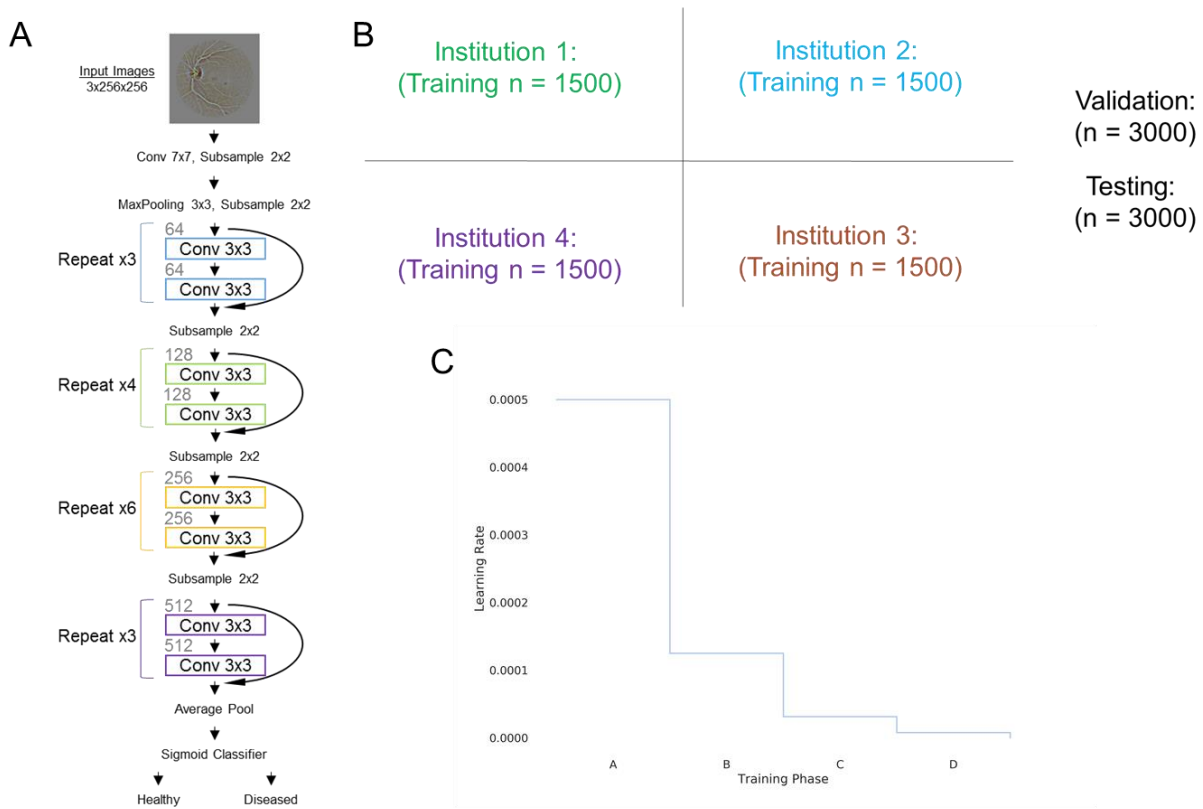

Figure 1. (A) ResNet-34 architecture was utilized for the Diabetic Retinopathy dataset. (B) The dataset was randomly divided into 4 institutions along with a validation and testing set. (C) The learning rate was decayed to .25 of its value when the same input samples are inputted into the network 20 times at a given learning rate without an improvement of the validation loss.

*Convolutional Neural Network*
We utilized the 34-layer residual network (ResNet34) architecture (Fig. 1A).[12] Our implementation was based on the Keras package with Theano backend.[13,14] The convolutional neural networks were run on a NVIDIA Tesla P100 GPU. During training, the probability of samples belonging to Healthy or Diseased class was computed with a sigmoid classifier. The weights of the network were optimized via a stochastic gradient descent algorithm with a mini-batch size of 32. The objective function used was binary cross-entropy. The learning rate was set to .0005 and momentum coefficient of .9. The learning rate was decayed to .25 of its value when the same samples were input into the network 20 times at a given learning rate with no improvement of the validation loss. The learning rate was decayed a total of 3 times (Training Phases A-D, Fig 1C). Biases were initialized using the Glorot uniform initializer.[15] To prevent overfitting and to improve learning, we augmented the data by introducing random rotations (0-360 degrees) and flips (50% change of horizontal or vertical) of the images at every epoch. Data augmentation was performed in real time in order to minimize memory usage. To further prevent overfitting, we utilized batch normalization after every convolutional layer.[16] The final model was evaluated by calculating the accuracy on the unseen testing cohort.

*Model Training Heuristics*

We tested several different training heuristics (Fig. 2) and compared the results. The first heuristic is training a neural network for each institution individually, assuming there is no collaboration between the institutions. The second heuristic is collaboration through pooling of all patient data into a shared dataset (centrally hosted data, Fig. 2A). The third heuristic was averaging the output of the four models trained on the institutions individually (ensemble single institution models, Fig. 2B). The fourth heuristic was training a model at a single institution until plateau of validation loss and then transferring the model to the next institution (single weight transfer, Fig. 2C). Under the single weight transfer training heuristic, the model is transferred to each institution exactly once. The last heuristic was training a model at each institution for a predetermined number of epochs (weight transfer frequency) before transferring the model to the next institution (cyclical weight transfer, Fig. 2D). Under the cyclical weight transfer training heuristic, the model is transferred to each institution more than once. The frequencies of weight transfer we studied were every 20 epochs, 10 epochs, 5 epochs, 4 epochs, 2 epochs, and every epoch.

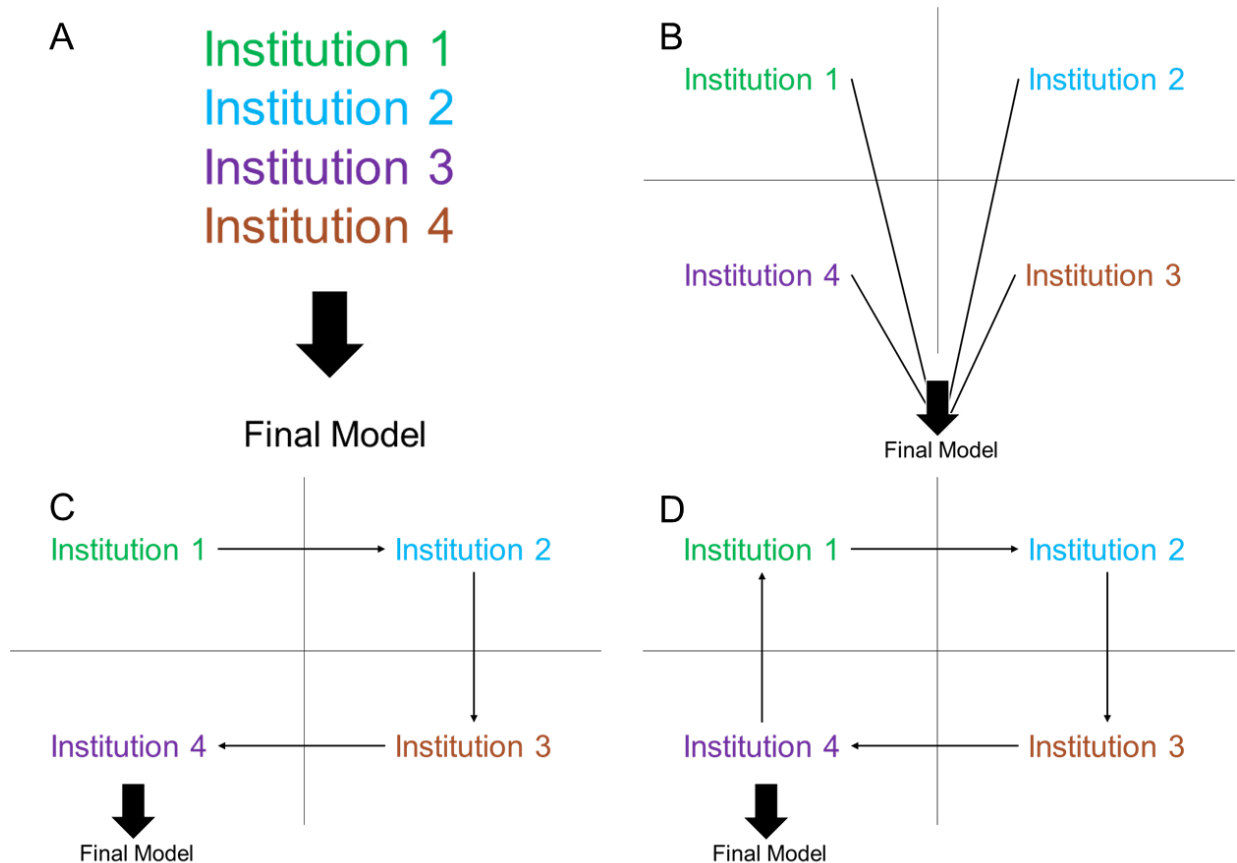

Figure 2. Model training heuristics investigated include (A) centrally hosted, (B) ensemble single institution models, (C) single weight transfer, and (D) cyclical weight transfer.

*Cyclical Weight Transfer With 20 Institutions*
We next addressed whether cyclical weight transfer can improve model performance when the performance of any individual institution is no better than random classification. To do this, we divided 6000 patient samples from the Kaggle Diabetic Retinopathy dataset into 20 institutions (n = 300 per institution) with equal class distributions. As with our previous experiments, we also sampled a single validation cohort (n = 3000 patient samples) and a single testing cohort (n = 3000 patient samples) with equal class probabilities. We then performed experiments with different numbers of collaborating institutions, starting with 1 and increasing to all 20 institutions. We utilized the cyclical weight transfer training heuristic with a weight transfer frequency of 1 epoch. We evaluated model performance via testing cohort accuracy. We compared testing accuracies with that of random classification and with the testing accuracy of a model trained with all 6000 patient samples centrally hosted.

*Repetition of Experiments in a Second Image Collection*
To demonstrate the reproducibility of our results, we repeated our experiments on Curated Breast Imaging Subset of the Digital Database for Screening Mammography (DDSM) dataset, an open source labeled dataset of mammograms.[17] For each patient, the dataset includes cranial-caudal and/or mediolateral-oblique views of the right and/or left breast, and each image is labeled as benign or malignant. For our experiments, we use a subset of 1508 grayscale images from 800 patients that had a mass in the breast. Along with each image, a binary segmentation mask for the mass was available. Of the 1508 images, 722 were labeled malignant and 786 were labeled benign, so a majority classifier would have 52.1% accuracy. We randomly selected 140 patients for each of the 4 "institutions", 120 patients for the validation cohort, and 120 patients for the testing cohort. This resulted in 257 images in Institution 1, 266 images in Institution 2, 257 images in Institution 3, and 270 images in Institution 4 (total of 1050 training images), 229 images in the validation set, and 229 images in the testing set. For the same patient, the different images, including different views of the same breast, could have different labels. Thus, we treated each image separately, but did not allow images from the same patient to be divided across different institutions, or across the training and testing/validation cohorts (as in our experiments with the Kaggle Diabetic Retinopathy dataset).

The grayscale image pixels were scaled between 0 and 1, and the mask pixels were either 0 or 1. Each image was cropped into 256x256 pixel resolution such that the region of interest as indicated by the binary mask was centered in the largest possible bounding box. Each cropped grayscale image along with its corresponding cropped binary mask were combined to produce a 2-channel 256x256 image. The images were normalized by subtracting the maximum pixel intensity and zero-centered by subtracting the mean pixel intensity. These normalized 2x256x256 images were input into a neural network. For this dataset, we used a 22-layer GoogLeNet with batch normalization after each convolutional layer, batch size of 32, and dropout of 0.5 before the final readout layer.[18] We used Adam optimizer with initial learning rate of 0.001 and learning rate decay of 0.99 every epoch (every 4 epochs in the weight transfer experiments) to optimize the model.[19] Cross entropy with L2 regularization coefficient of 0.0001 was used as the loss function. Model learning is terminated when there were 80 epochs of no improvement in validation loss (320 epochs in the weight transfer experiments). For the single weight transfer experiment, weights were transferred to the next institution each time there were 20 epochs of no improvement in validation loss, and learning was terminated when there were 20 epochs having no improvement in validation loss at the final institution. For ensembling, the output probabilities from the models trained at each of the 4 institutions were averaged to produce final class predictions. During training, the data were augmented by introducing random rotations (0-360 degrees) and flips (50% change of horizontal or vertical) to the images at every epoch.

*Repetition of Experiments in a Non-Medical Image Collection*
We further demonstrate the reproducibility of our results by repeating our experiments on the ImageNet dataset. We utilized the ImageNet 2012 classification dataset, which contains 1.28 million training images and 1000 classes.[20] To decrease the time of training, we utilized a subset of the training images for our experiments. We randomly selected 20 classes of the 1000 to work with. We randomly allocated 75 images of each class to each "institution" and 150 images of each class to the validation and testing cohort. In total, each of the 4 institutions had 1500 images and both the validation and testing cohorts had 3000 images. For pre-processing, we resized each image to 224x224 and subtracted by the per channel mean of the entire ImageNet dataset. As with the experiments with the Kaggle Diabetic Retinopathy dataset, we utilized the 34-layer residual network architecture. The learning rate was set to .0001 and momentum coefficient was set to .9. The learning rate was decayed to .25 of its value when the same samples were inputted into the network 20 times at a given learning rate with no improvement of the validation loss. To prevent overfitting and improve learning, we augmented the data by introducing random rotations (0-360 degrees), flips (50% change of horizontal or vertical), zooming (from -20% to +20%), and shearing (0 to .2 radians) at every epoch. We evaluated our models by assessing both the top-1 and top-5 accuracies. Top-1 accuracy is calculated by comparing the ground truth label with the top predicted class. Top-5 accuracy is calculated by comparing the ground truth label with the top 5 predicted classes.

## Results

### Diabetic Retinopathy Dataset

#### Single Institution Training
The models trained on single institutions had poor performance (Fig 3A-D). The average testing accuracies for the single institution models was 56.3% (Table 1). The highest testing accuracy for a network trained on a single institution was 59.0%.

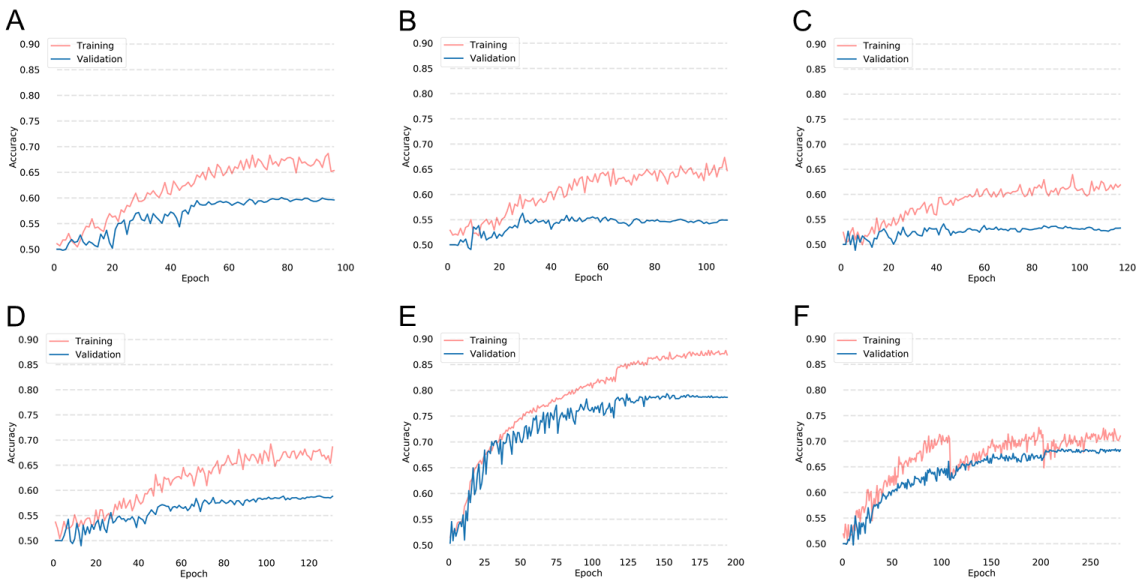

Figure 3. Performance of a neural network when trained on (A) Institution 1, (B) Institution 2, (C) Institution 3, and (D) Institution 4 for the Diabetic Retinopathy dataset. The training and validation accuracies for a model trained the centrally hosted training and single weight transfer training heuristics are shown in (E) and (F), respectively.

| Diabetic Retinopathy | Training Accuracy (n = 1500, %) | Validation Accuracy (n = 3000, %) | Testing Accuracy (n = 3000, %) |
|---|---|---|---|
| Institution 1 | 68.1 | 59.6 | 59.0 |
| Institution 2 | 66.8 | 54.9 | 53.8 |
| Institution 3 | 64.3 | 53.3 | 54.3 |
| Institution 4 | 69.5 | 58.8 | 58.2 |

| DDSM | Training Accuracy (n = 257-270, %) | Validation Accuracy (n = 229, %) | Testing Accuracy (n = 229, %) |
|---|---|---|---|
| Institution 1 | 59.1 | 55.5 | 55.0 |
| Institution 2 | 56.1 | 57.2 | 52.8 |
| Institution 3 | 59.0 | 52.8 | 60.3 |
| Institution 4 | 61.6 | 56.3 | 54.6 |

| ImageNet | Training Accuracy (n = 1500, %) | | Validation Accuracy (n = 3000, %) | | Testing Accuracy (n = 3000, %) | |
|---|---|---|---|---|---|---|
| | Top-1 | Top-5 | Top-1 | Top-5 | Top-1 | Top-5 |
| Institution 1 | 62.1 | 93.5 | 30.4 | 71.4 | 31.0 | 71.2 |
| Institution 2 | 66.1 | 95.0 | 31.1 | 70.0 | 32.4 | 71.5 |
| Institution 3 | 64.5 | 94.3 | 31.5 | 71.3 | 32.4 | 71.1 |
| Institution 4 | 66.8 | 94.5 | 31.6 | 70.8 | 32.1 | 71.6 |

Table 1. Training, validation, and testing accuracy of the neural network when trained on single institutions for the Diabetic Retinopathy, DDSM, and ImageNet datasets.

*Centrally Hosted Training*
When patient data from all institutions were pooled together, the collective size of the dataset was 6000. A network trained on the combined dataset had a high performance with a testing accuracy of 78.7% (Fig. 3E, Table 2).

| Diabetic Retinopathy | Training Accuracy (n = 6000, %) | Validation Accuracy (n = 3000, %) | Testing Accuracy (n = 3000, %) |
|---|---|---|---|
| Centrally Hosted | 89.4 | 78.6 | 78.7 |
| Ensemble Models | 63.2 | 60.9 | 60.0 |
| Single Weight Transfer | 70.4 | 68.3 | 68.1 |

| DDSM | Training Accuracy (n = 1050, %) | Validation Accuracy (n = 229, %) | Testing Accuracy (n = 229, %) |
|---|---|---|---|
| Centrally Hosted | 77.0 | 71.6 | 70.7 |
| Ensemble Models | 63.7 | 56.3 | 61.1 |
| Single Weight Transfer | 61.3 ± 0.9 | 61.2 ± 0.8 | 61.1 ± 1.8 |

| ImageNet | Training Accuracy (n = 6000, %) | | Validation Accuracy (n = 3000, %) | | Testing Accuracy (n = 3000, %) | |
|---|---|---|---|---|---|---|
| | Top-1 | Top-5 | Top-1 | Top-5 | Top-1 | Top-5 |
| Centrally Hosted | 82.9 | 98.4 | 49.5 | 83.4 | 48.9 | 83.8 |
| Ensemble Models | 50.2 | 88.6 | 37.0 | 76.5 | 38.6 | 77.0 |
| Single Weight Transfer | 45.5 | 84.5 | 36.0 | 76.2 | 37.9 | 75.5 |

Table 2. Training, validation, and testing accuracy of centrally hosted training, ensembling single institution model outputs, and single weight transfer for for Diabetic Retinopathy, DDSM, and ImageNet datasets.

*Ensembling Single Institution Models*
Averaging the sigmoid probability of the single institution models resulted in a testing accuracy of 60.0% (Table 2). Notably, the ensembled model outperformed any network trained on a single institution in terms of validation and testing accuracy.

*Single Weight Transfer*
Using single weight transfer heuristic, the model was trained at each institution until the plateau of validation loss was reached, followed by transferring of the model to the next institution. The resulting model had a testing accuracy of 68.1% (Fig. 3F, Table 2).

*Cyclical Weight Transfer*
Using a cyclical weight transfer heuristic, the model was transferred to the next institution at a prespecified frequency. In our initial experiment, we trained the network for 20 epochs at each institution before transferring the weights to the next institution. The average testing accuracy after repeating this experiment 3 times was 76.1% (Fig. 4A, Table 3).

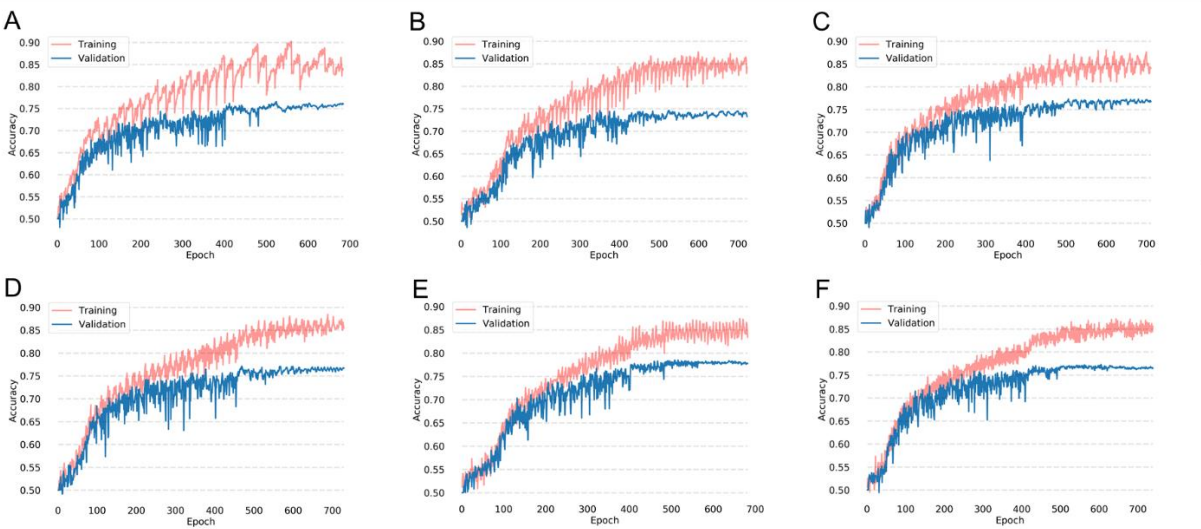

Figure 4. Training and validation accuracies during training on the Diabetic Retinopathy dataset with cyclical weight transfer with weight transfer frequencies of every (A) 20 epochs, (B) 10 epochs, (C) 5 epochs, (D) 4 epochs, (E) 2 epochs, or (F) every epoch.

| Diabetic Retinopathy | Training Accuracy (n = 6000, %) | Validation Accuracy (n = 3000, %) | Testing Accuracy (n = 3000, %) |
|---|---|---|---|
| Cyclical Weight Transfer, Every: | | | |
| 20 Epochs | 85.8 ± 0.9 | 76.0 ± 0.6 | 76.1 ± 1.0 |
| 10 Epochs | 87.9 ± 1.6 | 75.6 ± 2.0 | 75.9 ± 1.2 |
| 5 Epochs | 86.8 ± 0.9 | 76.1 ± 0.6 | 76.1 ± 0.8 |
| 4 Epochs | 88.9 ± 1.1 | 76.6 ± 0.1 | 77.4 ± 0.2 |
| 2 Epochs | 89.1 ± 1.7 | 77.3 ± 0.5 | 77.8 ± 0.3 |
| Epoch | 89.4 ± 2.3 | 77.3 ± 1.3 | 77.3 ± 0.9 |
| DDSM | Training Accuracy (n = 1050, %) | Validation Accuracy (n = 229, %) | Testing Accuracy (n = 229, %) |

| Cyclical Weight Transfer, Every: | | | |
|---|---|---|---|
| 20 Epochs | 72.7 ± 1.3 | 66.5 ± 3.5 | 65.4 ± 1.1 |
| 10 Epochs | 70.5 ± 4.7 | 68.9 ± 0.9 | 68.1 ± 3.6 |
| 5 Epochs | 71.5 ± 3.0 | 69.1 ± 0.2 | 68.1 ± 1.2 |
| 4 Epochs | 71.7 ± 1.9 | 65.9 ± 1.8 | 68.7 ± 2.4 |
| 2 Epochs | 71.9 ± 1.5 | 69.3 ± 2.4 | 69.9 ± 2.7 |
| Epoch | 74.8 ± 2.0 | 68.9 ± 1.3 | 69.1 ± 2.9 |

| ImageNet | Training Accuracy (n = 6000, %) | | Validation Accuracy (n = 3000, %) | | Testing Accuracy (n = 3000, %) | |
|---|---|---|---|---|---|---|
| | Top-1 | Top-5 | Top-1 | Top-5 | Top-1 | Top-5 |
| Cyclical Weight Transfer, Every: | | | | | | |
| 20 Epochs | 77.2 ± 3.2 | 97.7 ± 0.8 | 46.9 ± 0.8 | 82.8 ± 0.7 | 46.6 ± 0.9 | 83.2 ± 0.9 |
| 10 Epochs | 78.5 ± 1.2 | 98.0 ± 0.4 | 47.8 ± 0.9 | 82.9 ± 0.4 | 47.3 ± 0.6 | 83.8 ± 0.1 |
| 5 Epochs | 77.7 ± 2.6 | 97.7 ± 0.4 | 47.7 ± 0.7 | 83.0 ± 0.1 | 47.5 ± 1.4 | 83.3 ± 0.5 |
| 4 Epochs | 78.5 ± 3.5 | 97.9 ± 0.6 | 47.2 ± 0.9 | 83.2 ± 0.5 | 48.1 ± 0.6 | 83.6 ± 0.2 |
| 2 Epochs | 79.0 ± 3.2 | 97.8 ± 0.9 | 47.9 ± 0.0 | 82.8 ± 0.4 | 47.6 ± 1.1 | 84.1 ± 0.4 |
| Epoch | 83.2 ± 3.5 | 98.6 ± 0.6 | 49.2 ± 0.3 | 83.9 ± 0.7 | 49.3 ± 1.0 | 84.7 ± 0.1 |

Table 3. Training, validation, and testing accuracy for cyclical weight transfer for Diabetic Retinopathy, DDSM, and ImageNet datasets. Weight transfer frequencies investigated include every 20 epochs, 10 epochs, 5 epochs, 4 epochs, 2 epochs, and epoch. The accuracies for cyclical weight transfer are shown as mean ± standard deviation for 3 repetitions.

We also investigated whether having a higher frequency of weight transfer can improve the testing accuracy. We experimented with weight transfer frequencies of 10, 5, 4, 2, and every epoch, repeating each experiment 3 times (Fig. 4, Table 3). The average testing accuracy of lower frequency weight transfer (every 20, 10, or 5 epochs) was 76.1% while the average testing accuracy of higher frequency weight transfer (every 4, 2, or 1 epoch) was 77.5% (two-sample t-test p-value < .001). Thus, a higher frequency weight transfer had a statistically significant increase in testing accuracy. The average training testing accuracy for all cyclical weight transfer experiments was 76.8% (Fig. 5A).

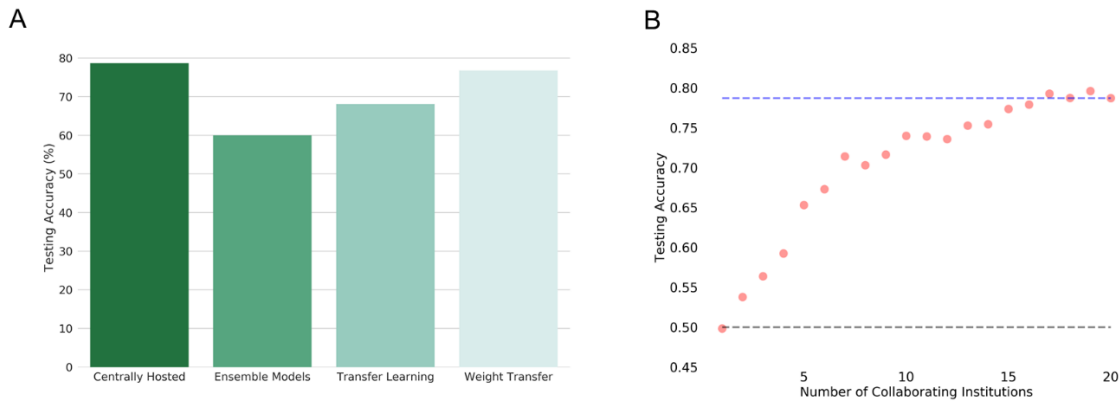

Figure 5. (A) Testing accuracies of centrally hosted training, ensembling models, single weight transfer, and cyclical weight transfer for our 4 "institution" experiment on the Diabetic Retinopathy dataset. Cyclical weight transfer had the performance that was on par with centrally hosted training. (B) To show distributed computation on a larger scale, we performed a 20 "institution" experiment with n = 300 patients per institution. The plot shown is the testing accuracy as a function of the number of collaborating institutions. All models were trained using the cyclical weight transfer training heuristic with a weight exchange frequency of 1. For reference, testing accuracy expected from random classification (gray line) and centrally hosted data (n = 6000 patients, blue line) are shown.

*Cyclical Weight Transfer With 20 Institutions*

We next addressed whether cyclical weight transfer can improve model performance when the performance of any individual institution is no better than random classification. To do this, we divided 6000 patient samples into 20 institutions, each with n = 300 patients. We trained models with increasing numbers of collaborating institutions, from 1 to 20. We utilized the cyclical weight transfer training heuristic with the weight transfer frequency of 1. As we increased the number of collaborating institutions, the testing accuracy increased (Fig. 5B). The testing accuracy for a single institution was 49.8%, which is equivalent to random classification as there are equal numbers of healthy and diseased patients. The testing accuracy for 20 collaborating institutions was 78.7%, which is on par with the performance of centrally hosted data with all 6000 patient samples.

*DDSM Dataset*
When we repeated the experiments on the DDSM dataset, the average testing accuracy was 55.7% for single institution models (Table 1, Supplemental Fig. 1A-D), only slightly better than a majority classifier. A model trained on centrally hosted data had a testing accuracy of 70.7% (Table 2, Supplemental Fig. 1E). Ensembling single institution models resulted in a testing accuracy of 61.1% and the single weight transfer training heuristic also resulted in an average testing accuracy of 61.1% (Table 2, Supplemental Fig. 1F). Cyclical weight transfer resulted in an average testing accuracy of 67.2% for low frequencies of weight transfer (every 20, 10, or 5 epochs), which was lower than the average testing accuracy of 69.2% for high frequency of weight transfer (every 4, 2, or 1 epoch, $p < .05$) (Supplemental Fig. 2, Table 3).

*ImageNet Dataset*
When these experiments were repeated for the ImageNet dataset, the average testing top-1 accuracy was 32.0% (top-5 accuracy = 71.4%) for single institution models (Table 1, Supplemental Fig. 3A-D). In comparison, a model trained on centrally hosted data had a testing top-1 accuracy of 48.9% (top-5 accuracy = 83.8%) (Table 2, Supplemental Fig. 3E). Ensembling single institution models resulted in a testing top-1 accuracy of 38.6% (top-5 accuracy = 77.0%), while the single weight transfer training heuristic resulted in a testing top-1 accuracy of 37.9% (top-5 accuracy = 75.5%) (Table 2, Supplemental Fig. 3F). Cyclical weight transfer resulted in an average testing top-1 accuracy of 47.1% (top-5 accuracy = 83.4%) for low frequencies of weight transfer (every 20, 10, or 5 epochs), which was lower than the average testing top-1 accuracy (48.3%, top-5 accuracy = 84.1%) for high frequency of weight transfer (every 4, 2, or 1 epoch, $p < .01$) (Table 3, Supplemental Fig. 4).

**Discussion**

All sharing heuristics, either data sharing or model sharing, outperformed models trained only on one institution in terms of testing accuracy. This shows the benefits of collaboration among multiple institutions in the context of deep learning. Unsurprisingly, a model trained on centrally hosted data had the highest testing accuracy, serving as a benchmark for the performance of our various model sharing heuristics.

To overcome limitations in data-sharing, we tried several approaches – ensembling of single institution models, single weight transfer, and cyclical weight transfer. Ensembling of neural networks trained to perform the same task is a common approach to significantly improve the generalization performance.[14] In comparison, the concept of single weight transfer is very similar to that of transfer learning, which is derived from that idea that a model can solve new problems faster by using knowledge learned from solving previous problems.[15] In practice, this involves training a model on one institution's dataset and fine-tuning the model on a different dataset. If we consider each institution as a separate dataset, the model is trained on institution 1 and fine-tuned on institutions 2, 3, and 4. Both ensembling single institution models and single weight transfer resulted in higher testing accuracies than any single institution model for Kaggle Diabetic Retinopathy, DDSM, and ImageNet datasets. Single weight transfer outperformed ensembling models for the Kaggle Diabetic Retinopathy dataset while ensembling models and single weight transfer had the same testing performance for the DDSM dataset. For the ImageNet dataset, ensembling models outperformed single weight transfer.

The highest testing accuracies amongst model sharing heuristics was cyclical weight transfer. On average, the testing accuracy of models trained with cyclical weight transfer was 1.9%, 2.5%, and 1.2% less than that of a model trained on centrally hosted data for the Kaggle Diabetic Retinopathy, DDSM, and ImageNet datasets, respectively. Furthermore, we find a higher frequency of weight transfer had a higher testing accuracy than a lower frequency of weight transfer. For the Kaggle Diabetic Retinopathy dataset, the higher frequency of weight transfer had, on average, a 1.4% increase in testing accuracy compared to lower frequency of weight transfer. Similarly, for the DDSM dataset, a higher frequency of weight transfer had, on average, a 2.0% increase in testing accuracy compared to lower frequency of weight transfer. Finally, for the ImageNet dataset, a higher frequency of weight transfer had, on average, a 1.1% increase in testing accuracy compared to lower frequency of weight transfer. The disadvantage of having a higher frequency of weight transfer, however, is that it may be more logistically challenging and may add to the total model training time. In these cases, a lower frequency of weight transfer would still produce results that are comparable to that of a model trained on centrally hosted data.

In our experiments with 4 institutions, we show that we are able to achieve high model performance without having the data centrally hosted. We next investigated whether high model performance can be achieved when the performance of any single institution is no better than random classification. We divided 6000 patient samples from the Diabetic Retinopathy dataset into 20 institutions, each with 300 patient samples. Indeed, when we trained a model using data from one institution, the performance was no better than random classification. As we increased the number of collaborating institutions (using cyclical weight transfer), we observed an increase in testing accuracy. With all 20 institutions, cyclical weight transfer achieved a testing accuracy on par with centrally hosted data with all 6000 patient samples. This simulates a scenario where patient data are distributed sparsely across many different institutions, and it is impossible to build a predictive model with data from any single institution. There are many situations (especially with rarer patient conditions) where no single institution has much patient data. In such cases, a distributed neural network can effectively utilize data from many institutions as long as the institutions are willing to share the model. In other words, if all institutions participate, they can, in essence, build a model capable of performing as if they had open access to all the data.

One of the limitations of our work was that each of the institutions was assumed to have the same number of patients and same class distribution. However, in real-world scenarios, these two conditions likely do not hold, and exploring the effect of imbalances in number of patients and class distribution on testing accuracy would be a topic of future investigation. Another limitation is that our "institutions" were sampled from a single dataset (such as Kaggle Diabetic Retinopathy dataset) and thus, do not display much variability from one institutions to the next. Future studies can explore scenarios where there is greater variability between institutions such as in the case where each institution is derived from a unique patient population. Lastly, we only investigated distributed learning in the context of a convolutional neural network. Distributed learning across institutions for other forms of deep learning, such as autoencoders, generative adversarial networks, and recurrent neural networks, warrant further study.

**Conclusion**

In this study, we address the question of how to train a deep learning model without sharing patient data. We found that cyclical weight transfer performed comparably to centrally hosted data, suggesting that sharing patient data may not always be necessary to build these models. We believe the results are generalizable with applications for many collaborative deep learning studies.

**Acknowledgements**

This project was supported by a training grant from the NIH Blueprint for Neuroscience Research (T90DA022759/R90DA023427). Its contents are solely the responsibility of the authors and do not necessarily represent the official views of the NIH.

This research was carried out in whole or in part at the Athinoula A. Martinos Center for Biomedical Imaging at the Massachusetts General Hospital, using resources provided by the Center for Functional Neuroimaging Technologies, P41EB015896, a P41 Biotechnology Resource Grant supported by the National Institute of Biomedical Imaging and Bioengineering (NIBIB), National Institutes of Health.

This study was also supported by National Institutes of Health grants U01 CA154601, U24 CA180927, U24 CA180918, U01 CA190214, and U01 CA187947.

We would like the acknowledge the GPU computing resources provided by the MGH and BWH Center for Clinical Data Science.

**Supplemental Data**

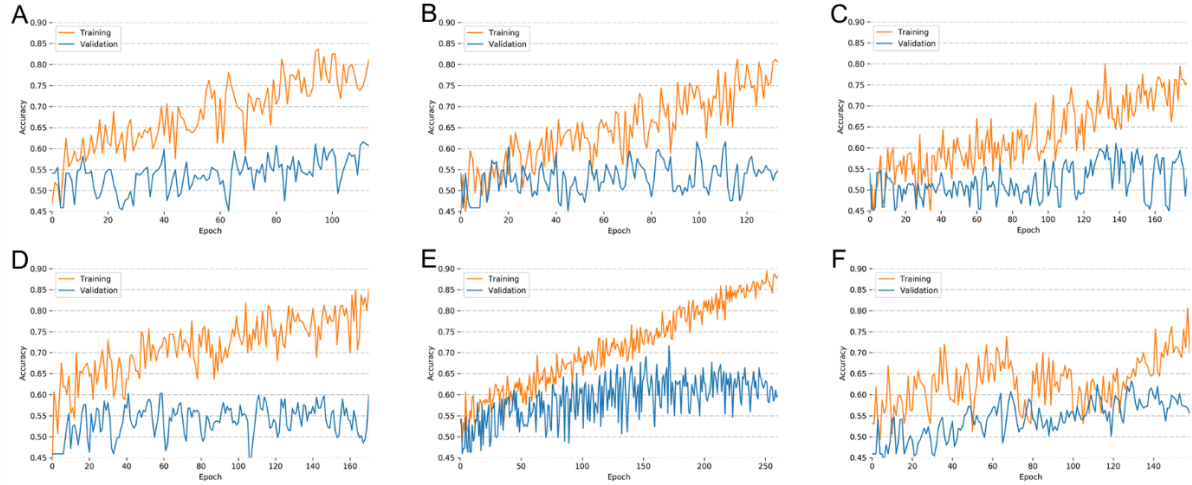

Supplemental Figure S1. Training and validation accuracies during training on DDSM dataset when trained on (A) Institution 1, (B) Institution 2, (C) Institution 3, and (D) Institution 4, (E) Centrally Hosted Training Heuristic, and (F) Single Weight Transfer Training Heuristic.

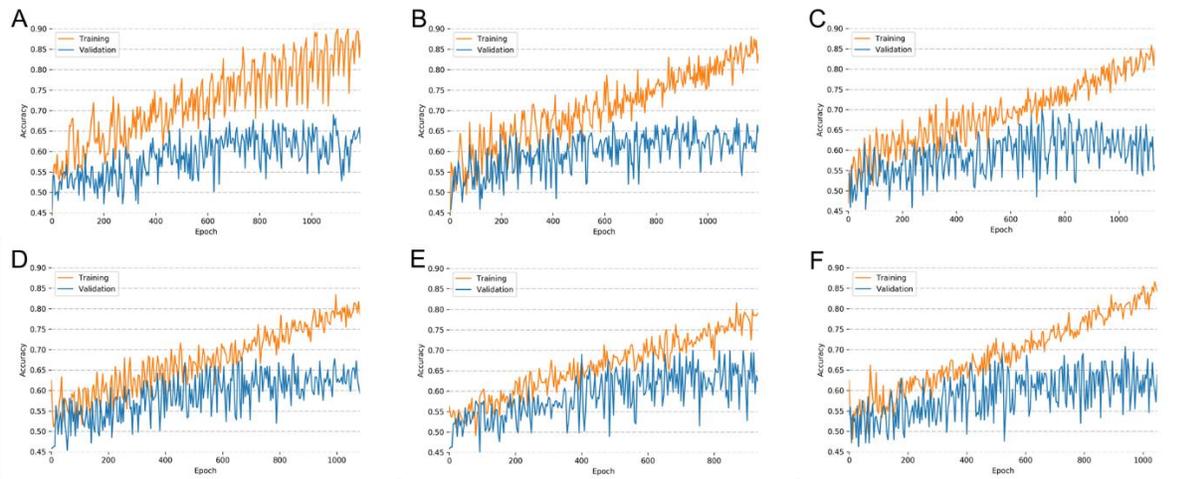

Supplemental Figure S2. Training and validation accuracies during training on the DDSM dataset with cyclical weight transfer with weight transfer frequencies of every (A) 20 epochs, (B) 10 epochs, (C) 5 epochs, (D) 4 epochs, (E) 2 epochs, or (F) every epoch.

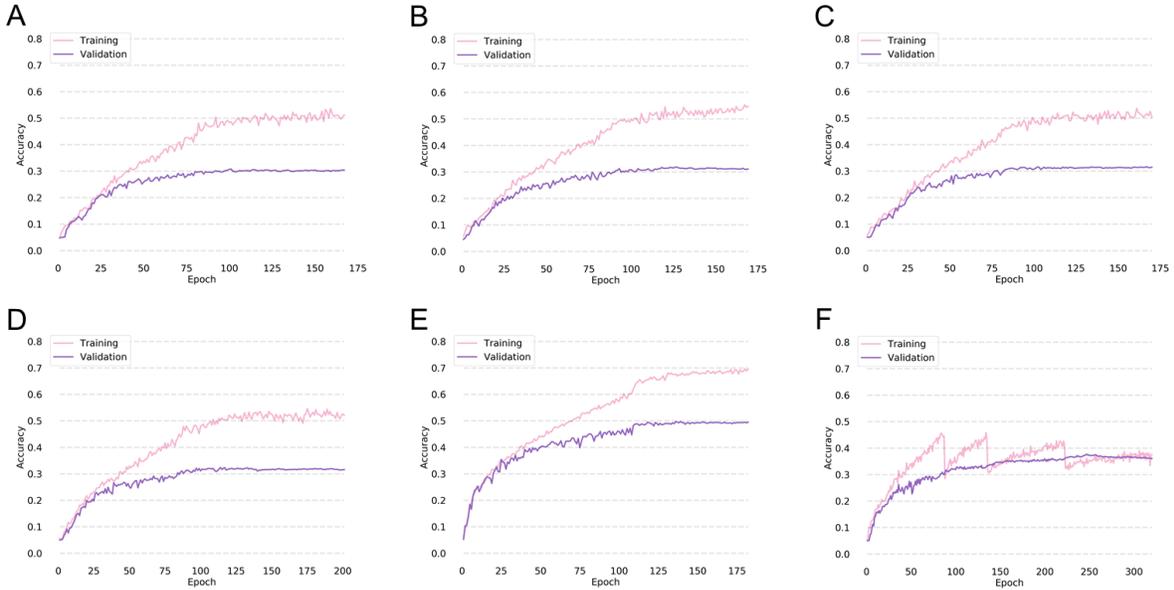

Supplemental Figure S3. Training and validation accuracies during training on ImageNet dataset when trained on (A) Institution 1, (B) Institution 2, (C) Institution 3, and (D) Institution 4, (E) Centrally Hosted Training Heuristic, and (F) Single Weight Transfer Training Heuristic.

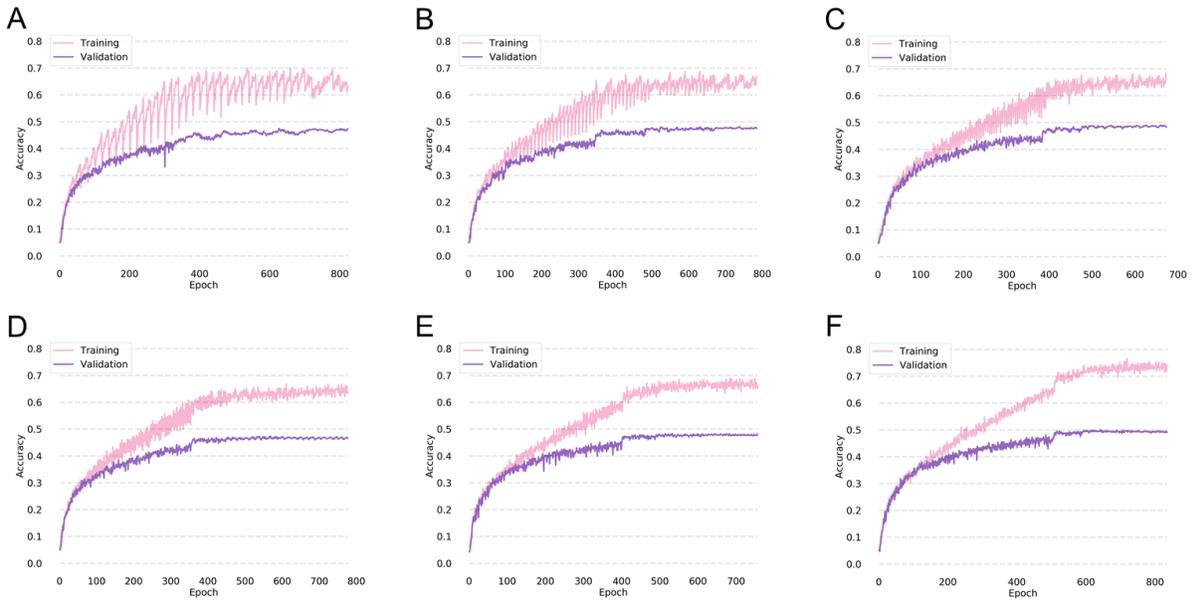

Supplemental Figure S4. Training and validation accuracies during training on the ImageNet dataset with cyclical weight transfer with weight transfer frequencies of every (A) 20 epochs, (B) 10 epochs, (C) 5 epochs, (D) 4 epochs, (E) 2 epochs, or (F) every epoch.